\RequirePackage{fix-cm}
\documentclass[smallextended]{svjour3}       
\smartqed  
\usepackage{graphicx}
\usepackage{mathtools}
\graphicspath{{images/}}
\begin{document}

\title{ECG arrhythmia classification using a 2-D convolutional neural network
}


\author{Tae Joon Jun\textsuperscript{1}		\and
        Hoang Minh Nguyen\textsuperscript{1}		\and
        Daeyoun Kang\textsuperscript{1}		\and
        Dohyeun Kim\textsuperscript{1}		\and
        Daeyoung Kim\textsuperscript{1}		\and
        Young-Hak Kim\textsuperscript{2}
}

\authorrunning{Tae Joon Jun et al} 

\institute{Tae Joon Jun \\ 
           taejoon89@kaist.ac.kr
           \\\\
           \textsuperscript{1} School of Computing, Korea Advanced Institute of Science and Technology, Daejeon, Republic of Korea\\
           \textsuperscript{2} Division of Cardiology, University of Ulsan College of Medicine, Asan Medical Center, Seoul, Republic of Korea\\
}

\date{Received: date / Accepted: date}

\maketitle

\begin{abstract}
In this paper, we propose an effective electrocardiogram (ECG) arrhythmia classification method using a deep two-dimensional convolutional neural network (CNN) which recently shows outstanding performance in the field of pattern recognition. Every ECG beat was transformed into a two-dimensional grayscale image as an input data for the CNN classifier. Optimization of the proposed CNN classifier includes various deep learning techniques such as batch normalization, data augmentation, Xavier initialization, and dropout. In addition, we compared our proposed classifier with two well-known CNN models; AlexNet and VGGNet. ECG recordings from the MIT-BIH arrhythmia database were used for the evaluation of the classifier. As a result, our classifier achieved 99.05\% average accuracy with 97.85\% average sensitivity. To precisely validate our CNN classifier, 10-fold cross-validation was performed at the evaluation which involves every ECG recording as a test data. Our experimental results have successfully validated that the proposed CNN classifier with the transformed ECG images can achieve excellent classification accuracy without any manual pre-processing of the ECG signals such as noise filtering, feature extraction, and feature reduction.
\end{abstract}

\section{Introduction}

According to the World Health Organization (WHO), cardiovascular diseases (CVDs) are the number one cause of death today. Over 17.7 million people died from CVDs which is an about 31\% of all deaths, and over 75\% of these deaths occur in low- and middle-income countries.\cite{whocvd} Arrhythmia is a representative type of CVDs that refers to any irregular change from the normal heart rhythms. There are several types of arrhythmia including atrial fibrillation, premature contraction, ventricular fibrillation, and tachycardia. Although single arrhythmia heartbeat may not have a serious impact on life, continuous arrhythmia beats can result in fatal circumstances. For example, prolonged premature ventricular contractions (PVCs) beats occasionally turn into a ventricular tachycardia (VT) or a ventricular fibrillation (VF) beats which can immediately lead to the heart failure. Thus, it is important to periodically monitor the heart rhythms to manage and prevent the CVDs. Electrocardiogram (ECG) is a non-invasive medical tool that displays the rhythm and status of the heart. Therefore, automatic detection of irregular heart rhythms from ECG signals is a significant task in the field of cardiology. 

Several methods have been presented in the literature for ECG arrhythmia classification. Firstly, we summarized the literature that used the feed-forward neural networks (FFNN) as the classifier. Linh et al\cite{linh2003line} presented fuzzy neural network model which applied Hermite function for the feature extraction. G\"uler et al\cite{guler2005ecg} also proposed FFNN as the classifier and WT for the feature extraction which used Lavenberg-Marquard algorithm for the training. Ceylan et al\cite{ceylan2007comparison} proposed FFNN as the classifier, principal component analysis (PCA) and wavelet transform (WT) for the feature extractions, and fuzzy c-means clustering (FCM) method for the feature reduction. As a result, they concluded that combination of PCA, FCM, and FFNN achieved the best result with 10 different arrhythmias. Detection of four different arrhythmias with average 96.95\% accuracy was obtained. Haseena et al\cite{haseena2011fuzzy} introduced FCM based probabilistic neural network (PNN) and achieved near 99\% average accuracy with eight different types of arrhythmia. Besides above literatures, \cite{yu2007electrocardiogram,yu2008integration,rai2013ecg,jadhav2010ecg,jadhav2010artificial,jun2016premature} also applied different FFNNs as the classifier with various feature extraction and feature elimination methods. 

Support vector machine (SVM) is also widely applied classification method in ECG arrhythmia detection. Osowski et al\cite{osowski2004support} introduced SVM as the classifier which used higher order statistics (HOS) and Hermite function as the feature extraction. Song et al\cite{song2005support} presented a combination of linear discriminant analysis (LDA) with SVM for six types of arrhythmia. Polat and G\"unes\cite{polat2007detection} proposed least square SVM (LS-SVM) with PCA reducing 279 features into 15 features. Melgani and Bazi\cite{melgani2008classification} proposed SVM with particle swarm optimization (PSO) for the classifier and compared their result with K-nearest neighbor (K-NN) and radial basis function (RBF) neural network classifiers. As a result, they achieved 89.72\% overall accuracy with six different arrhythmias.  Dutta et al\cite{dutta2010correlation} introduced cross-correlation based feature extraction method with LS-SVM classifier. Deasi et al\cite{desai2015machine} proposed five different arrhythmia detection with SVM classifier using discrete wavelet transform (DWT) for the feature extraction and independent component analysis (ICA) as the feature reduction method. Besides above literatures, \cite{kohli2010svm,ye2010arrhythmia,nasiri2009ecg} also applied SVM classifier as the ECG arrhythmia detection.

Various machine learning techniques are also used as the classifier besides FFNN and SVM. \"Ubeyli\cite{ubeyli2009combining} proposed recurrent neural networks (RNN)  classifier with eigenvector based feature extraction method. As a result, the model achieved 98.06\% average accuracy with four different arrhythmias. Kumar and Kumaraswamy\cite{kumar2012investigating} introduced random forest tree (RFT) as the classifier which used only RR interval as a classification feature. K-nearest neighbor (K-NN) is also popular classifier for the arrhythmia classification. Park et al\cite{park2013arrhythmia} proposed K-NN classifier for detecting 17 types of ECG beats which result in average 97.1\% sensitivity and 98.9\% specificity. Jun et al\cite{jun2016gpu} also introduced K-NN which proposed parallel K-NN classifier for the high-speed arrhythmia detection. Related to this paper, Kiranyaz et al\cite{kiranyaz2016real} introduced one-dimensional CNN for the ECG arrhythmia classification. The similarity exists in the use of CNN classifiers, but unlike our method of applying CNN to two-dimensional ECG images, Kiranyaz's method applied CNN to one-dimensional ECG signals, and our method is superior in performance. Rajpurkar et al\cite{rajpurkar2017cardiologist} also proposed a 1-D CNN classifier that used deeper and more data than the CNN model of Kiranyaz. Despite using a larger ECG dataset, the arrhythmia diagnostic performance is much lower than the literature. The reason is that although the size of the dataset is increased, the ECG signal used as an input remains in one dimension, so the performance improvement is not great even with deep CNN.

Although a number of the literature were proposed for the ECG arrhythmia classification, they have one or more of the following limitations: 1) good performance on carefully selected ECG recordings without cross-validation, 2) ECG beat loss in noise filtering and feature extraction schemes, 3) limited number of ECG arrhythmia types for the classification, 4) relatively low classification performance to adopt in practical.

In this paper, we firstly propose an ECG arrhythmia classification method using deep two-dimensional CNN with grayscale ECG images. By transforming one-dimensional ECG signals into two-dimensional ECG images, noise filtering and feature extraction are no longer required. This is important since some of ECG beats are ignored in noise filtering and feature extraction. In addition, training data can be enlarged by augmenting the ECG images which result in higher classification accuracy. Data augmentation is hard to be applied in previous literature since the distortion of one-dimensional ECG signal could downgrade the performance of the classifier. However, augmenting two-dimensional ECG images with different cropping methods helps the CNN model to train with different viewpoints of the single ECG images. Using ECG image as an input data of the ECG arrhythmia classification also benefits in the sense of robustness. Current ECG arrhythmia detection methods are sensitive to the noise signal since every ECG one-dimensional signal value is treated to have an equal degree of the classification. However, when the ECG signal is converted to the two-dimensional image, proposed CNN model can automatically ignore the noise data while extracting the relevant feature map throughout the convolutional and pooling layer. Thus, proposed CNN model can be applied to the ECG signals from the various ECG devices with different sampling rates and amplitudes while previous literature requires a different model for the different ECG devices. Furthermore, detecting ECG arrhythmia with ECG images resembles how medical experts diagnose arrhythmia since they observe an ECG graph from the patients throughout the monitor, which shows a series of ECG images. In other words, the proposed scheme can be applied to the medical robot that can monitors the ECG signals and helps the experts to identify ECG arrhythmia more precisely.

Our classification method consists the following steps: data acquisition, ECG data pre-processing, and CNN classifier. ECG signal data treated in this paper is obtained from the MIT-BIH database which is generally used as an arrhythmia database in ECG arrhythmia classification research. With these ECG recordings, we transformed every single ECG beat into 128 x 128 grayscale image since our CNN model requires two-dimensional image as an input. Different from the previous ECG arrhythmia classification works, our input data does not need to be separated into an exact single beat. In other words, even though additional signals from previous and afterward beats are in the image, our CNN model can automatically ignore those noise data when learning the model. Finally, the CNN classifier is optimized to classify eight different types of ECG beats as follows: normal beat (NOR), premature ventricular contraction beat (PVC), paced beat (PAB), right bundle branch block beat (RBB), left bundle branch block beat (LBB), atrial premature contraction beat (APC), ventricular flutter wave beat (VFW), and ventricular escape beat (VEB). Since the majority of the ECG signals in the MIT-BIH database is the normal beat, we augmented seven other arrhythmia beats by cropping the image in nine different ways at the training phase. By using the proposed augmentation, our model achieved more than 5\% higher weighted average sensitivity in the three arrhythmia types, premature contraction beat, ventricular flutter wave, and ventricular escape beat, without any modification in CNN model. In addition to data augmentation, we optimized our CNN model with modern deep learning techniques such as batch normalization\cite{ioffe2015batch}, dropout\cite{srivastava2014dropout}, and Xavier initialization\cite{glorot2010understanding}. The performance of the proposed classifier was evaluated with TensorFlow with the support of high-end NVIDIA GPUs. As a result, our CNN classifier achieved 99.05\% average accuracy, 99.57\% specificity, 97.85\% average sensitivity, and 98.55\% average positive predictive value while 10-fold cross validation method is applied to the evaluation to precisely validate the proposed classifier which involves every ECG recording as a test data.

The rest of this paper is organized as follows. Section 2 explains the detailed methodologies used for the ECG arrhythmia classification including ECG data pre-processing, and convolutional neural network classifier. Evaluation and experimental results of ECG arrhythmia classification are in Section 3. Finally, Section 4 draws discussion and conclusion of the paper.


\section{Methods}
\begin{figure}[!htb]
	\centering
    \includegraphics[width=\textwidth]{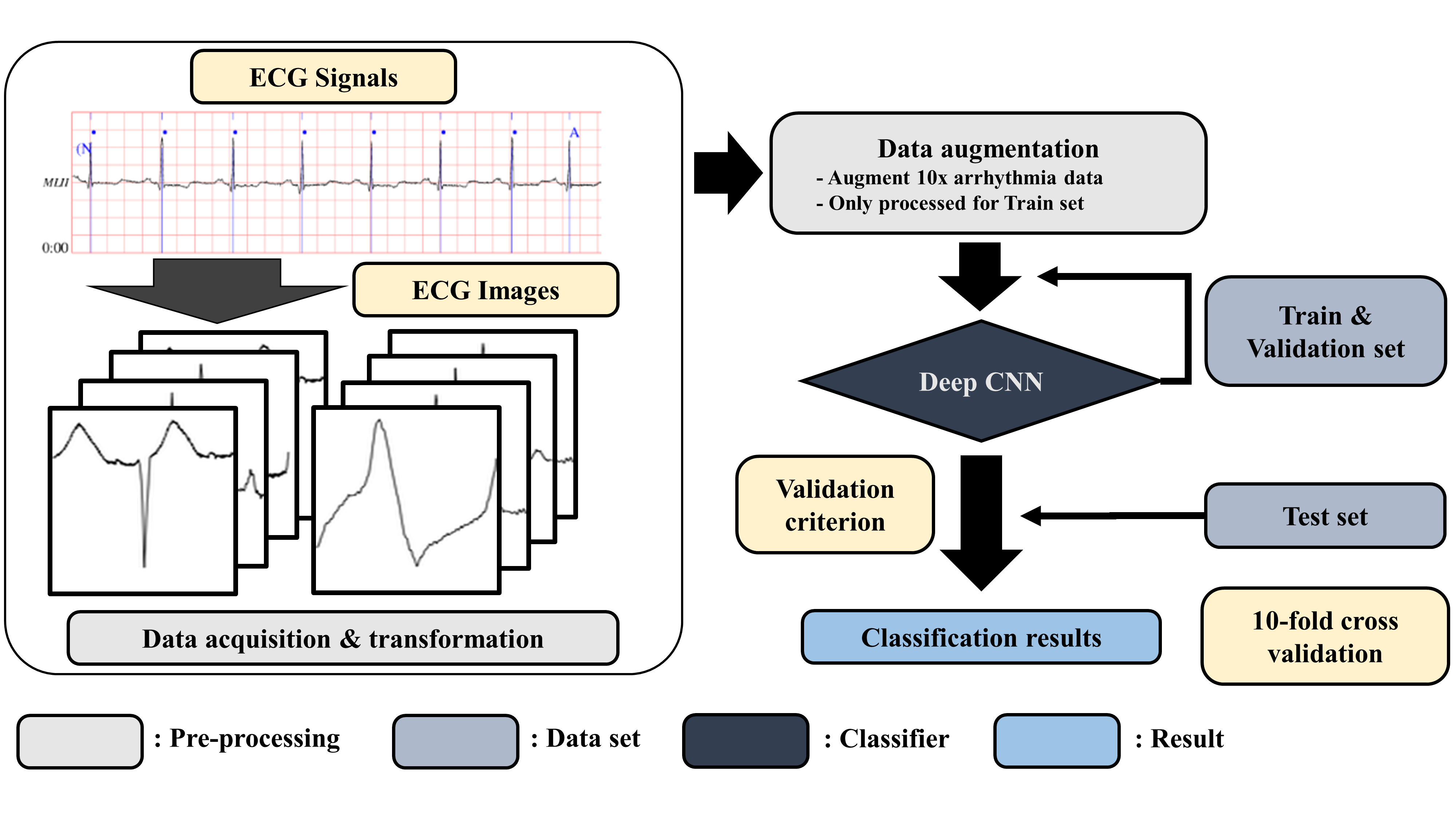}
    \caption{Overall procedures processed in ECG arrhythmia classification}
    \label{Fig1}
\end{figure}

Our CNN based ECG arrhythmia classification consists following steps: ECG data pre-processing, and the ECG arrhythmia classifier. In this paper, we used the MIT-BIH arrhythmia database\cite{moody2001impact} for the CNN model training and testing. Since the CNN model handles two-dimensional image as an input data, ECG signals are transformed into ECG images during the ECG data pre-processing step. With these obtained ECG images,
classification of eight ECG types is performed in CNN classifier step. Overall procedures are shown in Fig.\ref{Fig1}

\subsection{ECG data pre-processing}
Two-dimensional CNN requires image as an input data. Therefore, we transformed ECG signals into ECG images by plotting each ECG beat as an individual 128 x 128 grayscale image. In the MIT-BIH arrhythmia database, every ECG beat is sliced based on Q-wave peak time. More specifically, the type of arrhythmia is labeled at the Q-wave peak time of each ECG beat. Thus, we defined a single ECG beat image by centering the Q-wave peak signal while excluding the first and the last 20 ECG signals from the previous and afterward Q-wave peak signals. Based on the time information, a single ECG beat range can be defined in following:

\begin{equation}\label{(1)}
	\begin{split}
		T(Qpeak(n-1) + 20) \leq T(n) \leq T(Qpeak(n+1) - 20)
	\end{split}
\end{equation}

\begin{figure}[!htb]
	\centering
    \includegraphics[width=\textwidth]{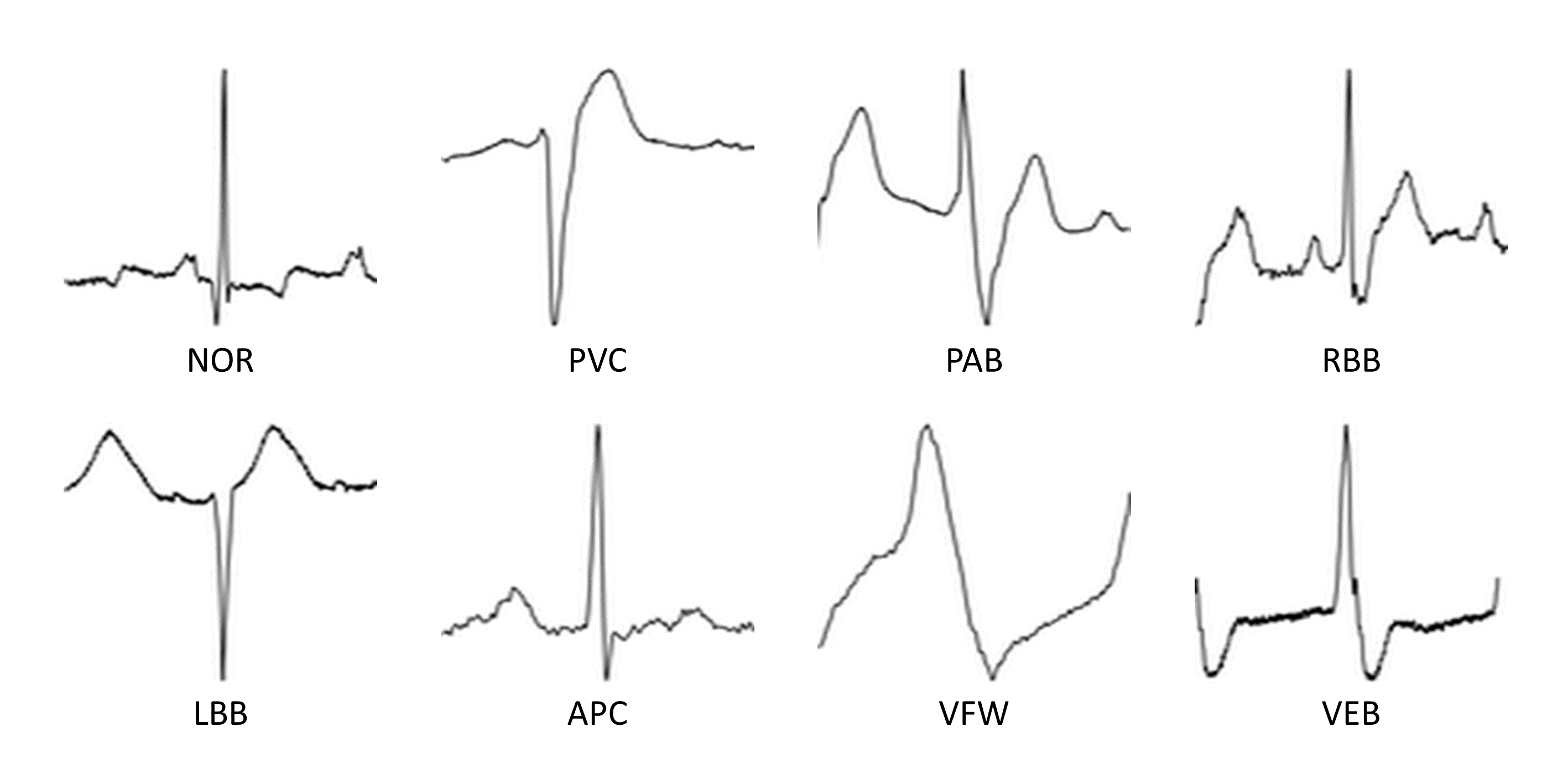}
    \caption{Normal beat and seven ECG arrhythmia beats}
    \label{Fig2}
\end{figure}

As a result, we obtained 100,000 images from the MIT-BIH arrhythmia database where each image is one of eight ECG beat types. Fig.\ref{Fig2} describes the eight ECG beat types with 128 x 128 grayscale images obtained from the ECG data pre-processing scheme.

\subsection{ECG arrhythmia classifier}
In this paper, we adopted CNN as the ECG arrhythmia classifier. CNN was first introduced by LeCun et al.\cite{lecun1989backpropagation} in 1989 and was developed through a project to recognize handwritten zip codes. The existing feed-forward neural network was not suitable for image classification since there is an exponential increase in the number of free parameters because the raw image is directly processed without considering the topology of the image. With the advent of the CNN model, correlation of spatially adjacent pixels can be extracted by applying a nonlinear filter and by applying multiple filters, it is possible to extract various local features of the image. The reason why we applied 2D CNN by converting the ECG signal into ECG image form in this paper is that 2D convolutional and pooling layers are more suitable for filtering the spatial locality of the ECG images. As a result, higher ECG arrhythmia classification accuracy can be obtained. In addition, the physician judges the arrhythmia in ECG signal of the patient through vision treatment through eyes. Therefore we concluded that applying the 2D CNN model to the ECG image is most similar to the physician's arrhythmia diagnosis process.

\begin{figure}[!htb]
	\centering
    \includegraphics[width=\textwidth]{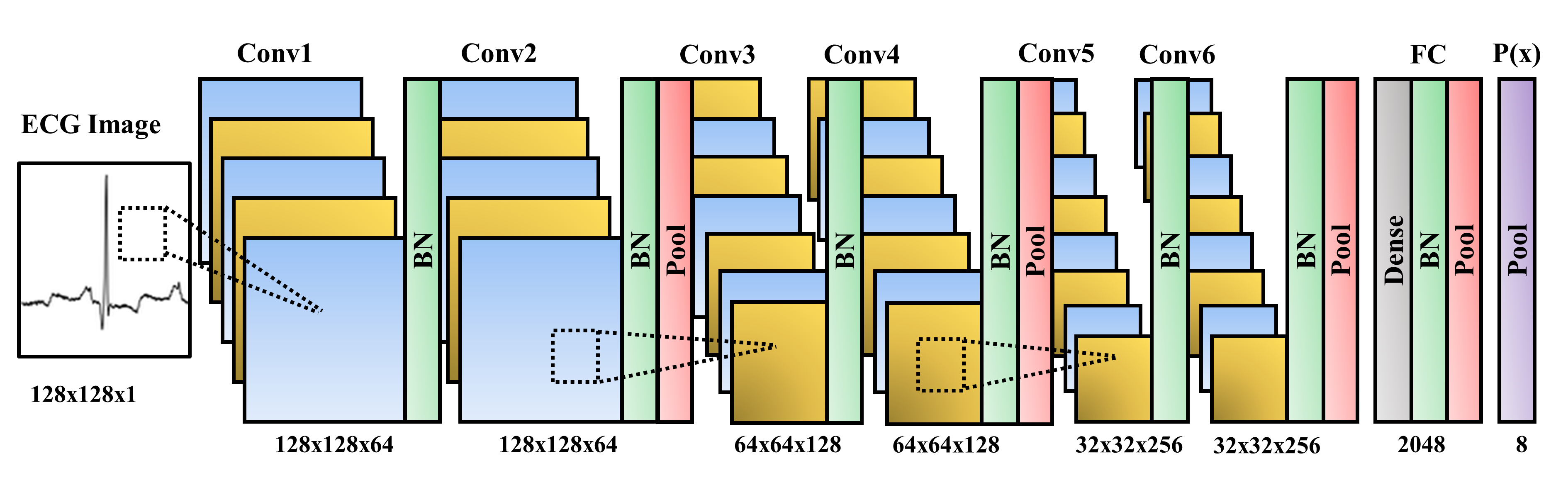}
    \caption{The architecture of proposed CNN model}
    \label{Fig3}
\end{figure}

There are several successful CNN models of the ImageNet Large Visual Perception Challenge (ILSVRC)\cite{russakovsky2015imagenet}, a competition for detecting and classifying objects in a given set of images. AlexNet\cite{krizhevsky2012imagenet}, announced in 2012, took first place with overwhelming performance as the first model to use the CNN model and GPU at ILSVRC. In 2014, GoogLeNet\cite{szegedy2015going} and VGGNet\cite{simonyan2014very} are the first and second in the ILSVRC, respectively. Although VGGNet is in second place, the VGGNet structure with repeating structure of 3x3 filters and subsampling is more often used in image recognition because the structure is much simpler and performance is not much different from GoogLeNet. ResNet\cite{he2016deep} and DenseNet\cite{huang2017densely}, which appeared in recent, were proposed to solve the problem that the effect of the early features of the image on the final output is small when the depth of the CNN model is deeper. In this paper, we follow the basic structure of VGGNet and optimize CNN model to show optimal performance for ECG arrhythmia classification. Performance comparison of the proposed CNN model was performed with AlexNet and VGGNet while did not compare with GoogLeNet, ResNet, and DenseNet. This is because the ECG image of this paper is a relatively simple 128 × 128 grayscale image, so there is no need to have a deep depth layer, and an increase in free parameters might cause over-fitting and degraded performance. Fig.\ref{Fig3} shows the overall architecture of proposed CNN model.\\ 

The following procedures are important optimization techniques that we considered while constructing the proposed CNN model.

\subsubsection{Data augmentation}
\begin{figure}[!htb]
	\centering
    \includegraphics[width=\textwidth]{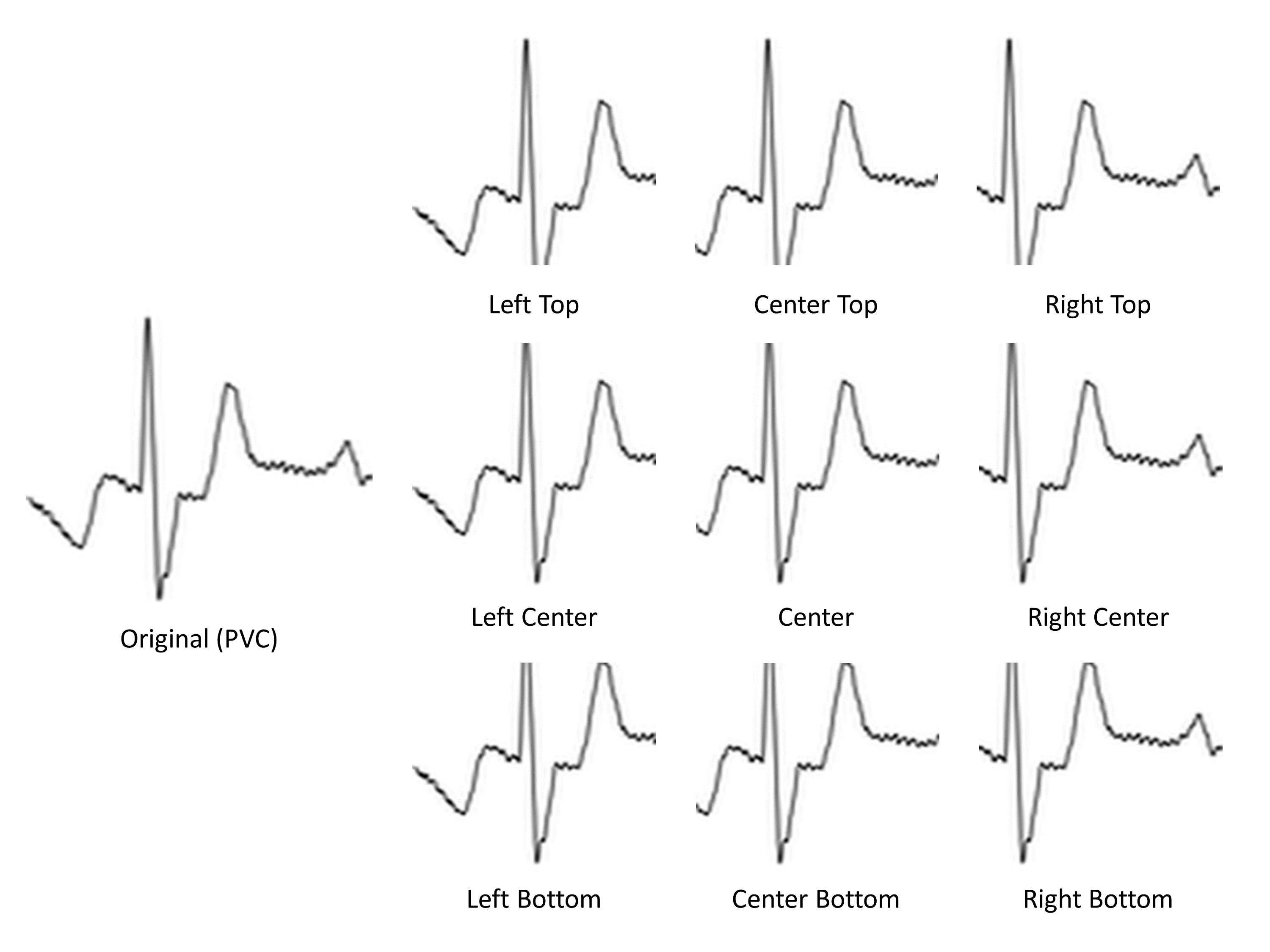}
    \caption{Original PVC image and nine cropped images}
    \label{Fig4}
\end{figure}

Data augmentation is one of the key benefits of using images as input data. The majority of previous ECG arrhythmia works could not manually add an augmented data into training set since the distortion of single ECG signal value may downgrade the performance in the test set. The reason is that unlike CNN, the other classifiers such as SVM, FFNN, and tree-based algorithm assumes that each ECG signal value has an equal worth of classification. However, since our CNN model defines an input data with two-dimensional ECG image, modifying the image with cropping and resizing does not downgrade the performance but enlarges the training data. When CNN is used as the classifier, data augmentation can effectively reduce overfitting and maintain a balanced distribution between classes. This advantage is particularly important in medical data analysis because most medical data are normal and only a few data are abnormal. In this case, due to the gradient descent learning, the normal loss corresponding to a plurality of classes in the batch is preferentially reduced, and a small class, which is an arrhythmia in this paper, is relatively ignored. As a result, while the specificity is very high in the training process, the sensitivity is likely to be low. In other words, we can achieve both high specificity and sensitivity by augmenting and balancing input data. In this work, we augmented seven ECG arrhythmia beats (PVC, PAB, RBB, LBB, APC, VFW, VEB) with nine different cropping methods: left top, center top, right top, center left, center, center right, left bottom, center bottom, and right bottom. Each cropping method results in two of three sizes of an ECG image, that is 96 x 96. Then, these augmented images are resized to the original size which is 128 x 128. Fig.\ref{Fig4} shows nine example cropped images with the original image with PVC. These augmented images are produced inside the model since copying these images from the disk increases the time spent on memory copy between the main memory and GPU memory, which result in slow learning speed.

\subsubsection{Kernel initialization}
The main pitfall of gradient descent based learning is that the model may diverge or fell into a local minimum point. Therefore, intelligent weight initialization is required to achieve convergence. In CNN, these weights are represented as kernels (or filters) and a group of kernels forms a single convolution layer. Our proposed CNN model uses the Xavier initialization. This initializer balances the scale of the gradients roughly the same in all kernels. The weights are randomly initialized in the following range:

\begin{equation}\label{(2)}
	x = \frac{\sqrt[]{6}}{In + Out}; [-x, x]
\end{equation}

Where \textit{In} and \textit{Out} are the number of input and output units at the kernel weights.

In other times, the CNN model usually starts with large size kernel with a small size depth and end with small size kernel with a large size depth. However, nowadays it is known that small size kernel with deeper layers is better than the previous approach. Therefore, we initialized our kernel size with 3 x 3 which is generally used in modern CNN classifier. We also include a zero-padding procedure to maintain the original size of the image after by-passing the convolution layer.

\subsubsection{Activation function}
The role of activation function is to define the output value of kernel weights in the model. In modern CNN models, nonlinear activation is widely used, including rectified linear units (ReLU), leakage rectified linear units (LReLU)\cite{maas2013rectifier}, and exponential linear units (ELU)\cite{clevert2015fast}. Although ReLU is the most widely used activation function in CNN, LReLU and ELU provide a small negative value because ReLU translates whole negative values to zero, which results in some nodes no longer participate in learning. After the experiment, we used ELU which showed better performance for ECG arrhythmia classification than LReLU. ReLU, LReLU, and ELU are shown in following:

\begin{equation}\label{(3)}
	ReLU(x) = max(0,x)
\end{equation}

\begin{equation}\label{(4)}
	LReLU(x) = max(0,x) + \alpha min(0,x)
\end{equation}

\begin{equation}\label{(5)}
	\begin{gathered}
		ELU(x) = \bigg\{
		\begin{aligned}
			x\hspace{2cm}if\hspace{0.2cm}x\geq0\\
			\gamma(exp(x)-1)\hspace{1cm}if\hspace{0.2cm}x<0
		\end{aligned}
	\end{gathered}
\end{equation}

Where the value of leakage coefficient ($\alpha$) is 0.3 and the value of the hyper-parameter($\gamma$) is 1.0.


\subsubsection{Regularization}
Regularization also called normalization, is a method to reduce the overfitting in the training phase. Typical normalization methods are L1 and L2 normalization, however, it is common to apply dropout and batch normalization in recent deep CNN models. In deep learning, when a layer is deepened, a small parameter change in the previous layer can have a large influence on the input distribution of the later layer. This phenomenon is referred to as internal covariate shift. Batch normalization has been proposed to reduce this internal covariate shift, and the mean and variance of input batches are calculated, normalized, and then scaled and shifted. The location of batch normalization is usually applied just before the activation function and after the convolution layer. However, from our experience, in some cases, it is better to place the batch normalization layer after the activation function, and the ECG arrhythmia classification is the case in this case. Therefore, we applied a batch normalization layer immediately after every activation function in the model, including the convolutional block and the fully-connected block. 

Dropout is a way to avoid overfitting with reducing the dependency between layers by participating nodes of the same layer probabilistically. In the training phase, Dropout does intentionally exclude some networks in learning, so the model can achieve the voting effect by the model combination. In this paper, we applied a dropout with a probability of 0.5 and placed its position after the batch-normalization layer of the fully-connected block. In general, Dropout is not applied to convolutional blocks because convolutional layers do not have many free-parameters and co-adaptation between nodes is more important than reducing the overfitting.

\subsubsection{Cost and optimizer function}

The cost function is a measure of how well the neural network is trained and represents the difference between the given training sample and the expected output. The cost function is minimized by using optimizer function. There are various types of cost functions, but deep learning typically uses a cross-entropy function.

\begin{equation}\label{(6)}
	C = - \frac{1}{n} \sum[y\ln a + (1-y)\ln (1-a)]
\end{equation}

Where \textit{n} is the number of training data (or the batch size), \textit{y} is an expected value, and \textit{a} is an actual value from the output layer.

To minimize the cost function, a gradient descent-based optimizer function with a learning rate is used. There are several well-known optimizer functions such as Adam\cite{kingma2014adam}, Adagrad\cite{duchi2011adaptive}, and Adadelta\cite{zeiler2012adadelta}. The final performance difference between the above functions was not large, but in our experiments, it was found that the optimal point reached the earliest when Adam was used. As result, we used the Adam optimizer function with an initial learning rate of 0.001 and 0.95 decay per 1,000 steps.

In our CNN model, we adopted Adam optimizer function with 0.0001 starting learning rate by exponentially decaying the learning rate every 1,000 decay steps with 0.95 decay rate. The learning rate at a given global step can be calculated as:

\begin{equation}\label{(7)}
	LearningRate = LearningRate_0 * 0.95^{\lceil(GlobalStep/1,000)\rceil}
\end{equation}

\subsubsection{Validation set}
The validation set is used to determine whether a model has reached sufficient accuracy with given training set. Without the validation procedure, the model is likely to fall overfitting. Generally, validation criterion for the CNN is the loss value. However, according to our observation, early stopping the model based on loss value could not achieve higher sensitivity in seven arrhythmia classification. Therefore, we set the average sensitivity of the validation set as the validation criterion. When there is no more increase in the weighted average sensitivity for the past 500 global steps, we stopped the learning procedure and starts the evaluation with the test set. 

\subsubsection{Optimized CNN classifier architecture}
\begin{figure}[!htb]
	\centering
    \includegraphics[width=\textwidth]{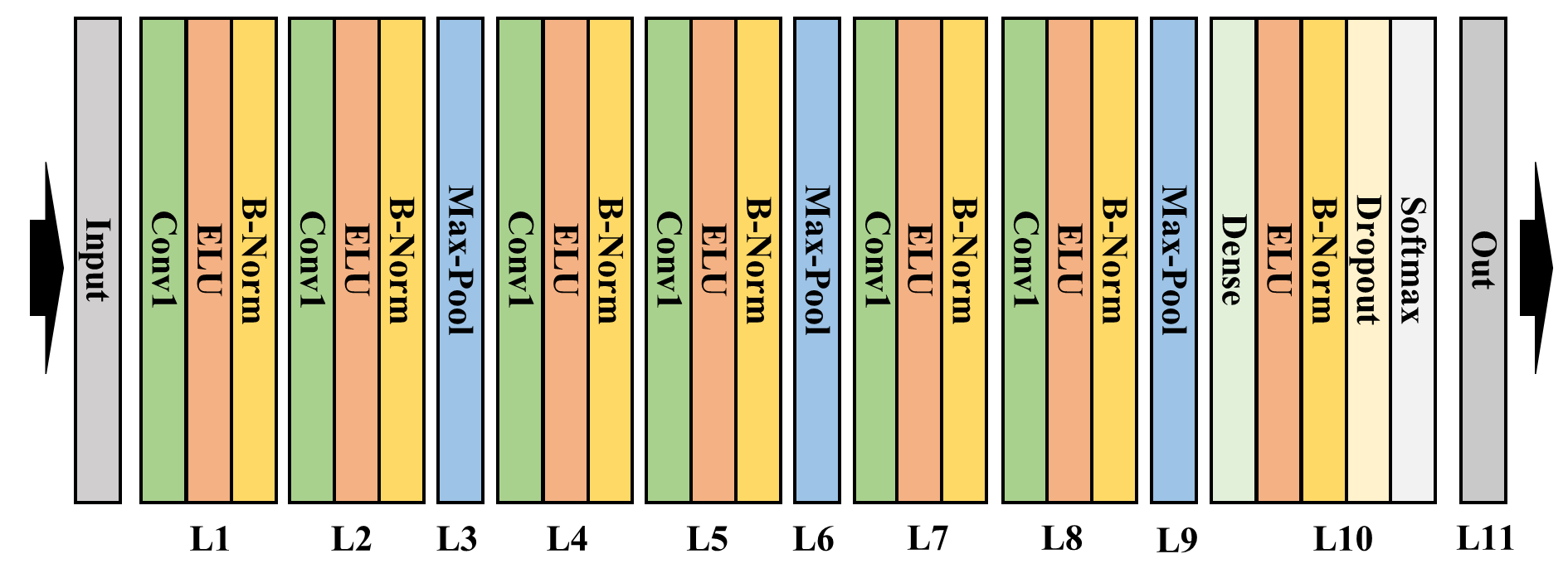}
    \caption{\textit{Architecture of proposed CNN Model}}
    \label{Fig5}
\end{figure}

Considering the above procedure, we designed the CNN model for the ECG arrhythmia classification. The main structure of the CNN model is similar to VGGNet, which optimizes various functions to reduce overfitting and improve classification accuracy. Table \ref{table1} and Fig.\ref{Fig5} describes detailed architecture of proposed CNN model. Since the proposed CNN model is compared to AlexNet and VGGNet, Table \ref{table2} and Table \ref{table3} represent the architectures of AlexNet and VGGNet we have deployed in the ECG arrhythmia classification.

\begin{table*}[!htb]\small
	\caption{Architecture of proposed CNN Model}
    \centering
    \scalebox{0.95}{
	\begin{tabular}{ lllllll}
    \hline\noalign{\smallskip}
		\textbf{        } & \textbf{    Type    } & \textbf{   Kernel size   } & \textbf{   Stride   } & \textbf{   \# Kernel   } & \textbf{   Input size   }\\
        \noalign{\smallskip}\hline\noalign{\smallskip}
		Layer1 & Conv2D & 3 x 3 & 1 & 64 & 128 x 128 x 1\\
		Layer2 & Conv2D & 3 x 3 & 1 & 64 & 128 x 128 x 64\\ 
        Layer3 & Pool & 2 x 2 & 2 &  & 128 x 128 x 64\\ 
        Layer4 & Conv2D & 3 x 3 & 1 & 128 & 64 x 64 x 64\\ 
        Layer5 & Conv2D & 3 x 3 & 1 & 128 & 64 x 64 x 128\\ 
        Layer6 & Pool & 2 x 2 & 2 &  & 64 x 64 x 128\\ 
        Layer7 & Conv2D & 3 x 3 & 1 & 256 &32 x 32 x 128 \\ 
        Layer8 & Conv2D & 3 x 3 & 1 & 256 & 32 x 32 x 256\\ 
        Layer9 & Pool & 2 x 2 & 2 &  & 32 x 32 x 256 \\ 
        Layer10 & Full &  & & 2048 & 16 x 16 x 256\\
        Layer11 & Out &  & & 8 & 2048\\
        \noalign{\smallskip}\hline
	\end{tabular}}
	\label{table1}
\end{table*}

\begin{table*}[!htb]\small
	\caption{Architecture of AlexNet }
    \centering
    \scalebox{0.95}{
	\begin{tabular}{ lllllll}
    \hline\noalign{\smallskip}
		\textbf{        } & \textbf{    Type    } & \textbf{   Kernel size   } & \textbf{   Stride   } & \textbf{   \# Kernel   } & \textbf{   Input size   }\\
        \noalign{\smallskip}\hline\noalign{\smallskip}
		Layer1 & Conv2D & 11 x 11 & 1 & 3 & 128 x 128 x 1\\
		Layer2 & Conv2D & 5 x 5 & 1 & 48 & 128 x 128 x 3\\ 
        Layer3 & Pool & 3 x 3 & 2 &  & 128 x 128 x 48\\ 
        Layer4 & Conv2D & 3 x 3 & 1 & 128 & 63 x 63 x 48\\ 
        Layer5 & Pool & 3 x 3 & 2 &  & 31 x 31 x 128\\ 
        Layer6 & Conv2D & 3 x 3 & 1 & 192 & 31x 31 x 128 \\ 
        Layer7 & Conv2D & 3 x 3 & 1 & 192 & 31 x 31 x 192\\
        Layer8 & Conv2D & 3 x 3 & 1 & 128 & 31 x 31 x 192\\ 
        Layer9 & Pool & 3 x 3 & 2 &  & 31 x 31 x 128 \\ 
        Layer10 & Full &  & & 2048 & 15 x 15 x 128\\
        Layer10 & Full &  & & 2048 & 2048\\
        Layer11 & Out &  & & 8 & 2048\\
        \noalign{\smallskip}\hline
	\end{tabular}}
    \scalebox{1}{
    \begin{tabular}{l}
    	cf) Batch normalization is not applied. Instead, local response normalization is \\used for convolution layers just before pooling layer.
    \end{tabular}}
	\label{table2}
\end{table*}

\begin{table*}[!htb]\small
	\caption{Architecture of VGGNet }
    \centering
    \scalebox{0.95}{
	\begin{tabular}{ llllll}
    \hline\noalign{\smallskip}
		\textbf{        } & \textbf{    Type    } & \textbf{   Kernel size   } & \textbf{   Stride   } & \textbf{   \# Kernel   } & \textbf{   Input size   }\\
        \noalign{\smallskip}\hline\noalign{\smallskip}
		Layer1 & Conv2D & 3 x 3 & 1 & 64 & 128 x 128 x 1\\
		Layer2 & Conv2D & 3 x 3 & 1 & 64 & 128 x 128 x 64\\ 
        Layer3 & Pool & 2 x 2 & 2 &  & 128 x 128 x 64\\ 
        Layer4 & Conv2D & 3 x 3 & 1 & 128 & 64 x 64 x 64\\ 
        Layer5 & Conv2D & 3 x 3 & 1 & 128 & 64 x 64 x 128\\ 
        Layer6 & Pool & 2 x 2 & 2 &  & 64x64x128\\ 
        Layer7 & Conv2D & 3 x 3 & 1 & 256 & 32 x 32 x 128 \\ 
        Layer8 & Conv2D & 3 x 3 & 1 & 256 & 32 x 32 x 256\\ 
        Layer9 & Conv2D & 3 x 3 & 1 & 256 & 32 x 32 x 256\\ 
        Layer10 & Pool & 2 x 2 & 2 &  & 32 x 32 x 256 \\
        Layer11 & Conv2D & 3 x 3 & 1 & 512 & 16 x 16 x 256 \\ 
        Layer12 & Conv2D & 3 x 3 & 1 & 512 & 16 x 16 x 512\\ 
        Layer13 & Conv2D & 3 x 3 & 1 & 512 & 16 x 16 x 512\\ 
        Layer14 & Pool & 2 x 2 & 2 &  & 8 x 8 x 512 \\
        Layer15 & Full &  & & 2048 & 8 x 8 x 512\\
        Layer16 & Full &  & & 2048 & 2048\\
        Layer17 & Out &  & & 8 & 2048\\
        \noalign{\smallskip}\hline
	\end{tabular}}
	\label{table3}
\end{table*}

\section{Experiments and Results}

The performance of the proposed CNN model was compared with two well known CNN models, AlexNet and VGGNet, and results of existing ECG arrhythmia classification literature. MIT-BIH arrhythmia database is used for the evaluation of the experiments. 


\subsection{Data acquisition}

The ECG arrhythmia recordings used in this paper are obtained from the MIT-BIH arrhythmia database. The database contains 48 half-hour ECG recordings collected from 47 patients between 1975 and 1979. The ECG recording is sampled at 360 samples per second. There are approximately 110,000 ECG beats in MIT-BIH database with 15 different types of arrhythmia including normal. The experiment of this paper is to validate the performance of our proposed CNN compared to that of well-known CNN models and previous ECG arrhythmia classification works. From the MIT-BIH database, we included normal beat (NOR) and seven types of ECG arrhythmias including premature ventricular contraction (PVC), paced beat (PAB), right bundle branch block beat (RBB), left bundle branch block beat (LBB), atrial premature contraction (APC), ventricular flutter wave (VFW), and ventricular escape beat (VEB). Table \ref{table4} describes the total ECG signals that we used from the MIT-BIH arrhythmia database. We excluded other seven types of arrhythmia such as a start of ventricular flutter beat, a fusion of paced and normal beat, and unclassifiable beat since these beat are generally ignored in ECG arrhythmia classification studies.

\begin{table}[!htb]\small
	\caption{ECG data description table }
    \centering
    \scalebox{1}{
	\begin{tabular}{ lll}
    \hline\noalign{\smallskip}
		\textbf{Type} & \textbf{Records} & \textbf{\#Beats}\\
        \noalign{\smallskip}\hline\noalign{\smallskip}
		NOR & 100,101,103,105,108,112,113,114,115,117,121,122,123 & 75052\\
        & 202,205,219,230,234 &\\
		PVC & 106,116,119,200,201,203,208,210,213,215,221,228,233 & 7130\\ 
        PAB & 102,104,107,217 & 7028\\ 
        RBB & 118,124,212,231 & 7259\\ 
        LBB & 109,111,207,213 & 8075\\ 
        APC & 209,220,222,223,232 & 2546\\ 
        VFW & 207 & 472\\ 
        VEB & 207 & 106\\
        \noalign{\smallskip}\hline\noalign{\smallskip}
        \textbf{Total} &   &  \textbf{106501}\\
        \noalign{\smallskip}\hline
	\end{tabular}}
	\label{table4}
\end{table}

\subsection{Experimental Setup}
Proposed classifier and two other CNN models are deployed in Python language with TensorFlow\cite{abadi2016tensorflow} which is an open source software library for deep learning launched by Google. Since CNN requires a lot of free parameters to train, GPGPU support is strongly recommended to reduce the learning time of the model. Thus, our experimental system is composed of two servers each contains two Intel Xeon E5 CPUs, 64GB main memory, and 2 NVIDIA K20m GPUs. With these NVIDIA GPUs, TensorFlow is accelerated by using CUDA and CUDNN\cite{chetlur2014cudnn}. Versions of each software are TensorFlow r1.0, CUDA 8.0, and CUDNN 5.5.

\subsection{ECG Arrhythmia classification evaluation}
\subsubsection{Evaluation metrics}
The evaluation of the classification considered five metrics: Area Under the Curve (AUC), Accuracy (Acc), Specificity (Sp), Sensitivity (Se), and Positive Predictive Value (+P). AUC is an area under the Receiver Operating Characteristic (ROC) curve. The AUC is calculated by Riemann sum with a set of thresholds to compute pairs of True Positive Rate (TPR) and False Positive Rate (FPR). Hosmer and Lemeshow provided the general guidelines for interpreting AUC values in \cite{hosmer2013applied}. Table \ref{table5} describes the brief guideline rules of the AUC interpretation. Specificity is the fraction of negative test results that are correctly identified as normal. Sensitivity is the probability of positive test results that are correctly identified as arrhythmia. Positive predictive value is the proportion of actual positive test result from the all positive calls. Except for the AUC, other four metrics are defined with four measurements in following:

\begin{table}[!htb]\small
	\caption{AUC interpretation guidelines}
    \centering
    \scalebox{1}{
	\begin{tabular}{ lll}
    \hline\noalign{\smallskip}
		\textbf{AUC} & \textbf{Guidelines} \\
        \noalign{\smallskip}\hline\noalign{\smallskip}
        0.5 - 0.6 & No discrimination\\ 
        0.6 - 0.7 & Poor discrimination\\ 
        0.7 - 0.8 & Acceptable discrimination\\
        0.8 - 0.9 & Good discrimination\\
        0.9 - 1 & Excellent discrimination\\
        \noalign{\smallskip}\hline
	\end{tabular}}
	\label{table5}
\end{table}

\begin{itemize}
\item True Positive (TP) : Correctly detected as arrhythmia
\item True Negative (TN) : Correctly detected as normal
\item False Positive (FP) : Incorrectly detected as arrhythmia
\item False Negative (FN) : Incorrectly detected as normal
\end{itemize}

The four metrics (Acc, Sp, Se, PPV) are defined in following:

\begin{equation}\label{(8)}
	Accuracy(\%) = \frac{TP + TN}{TP + TN + FP + FN} \times 100
\end{equation}

\begin{equation}\label{(9)}
	Specificity(\%) = \frac{TN}{FP + TN} \times 100
\end{equation}

\begin{equation}\label{(10)}
	Sensitivity(\%) = \frac{TP}{TP + FN} \times 100
\end{equation}

\begin{equation}\label{(11)}
	Positive Predictive Value(\%) = \frac{TP}{TP + FP} \times 100
\end{equation}

\subsubsection{\textit{K}-fold cross validation}
\begin{figure}[!htb]
	\centering
    \includegraphics[width=0.9\textwidth]{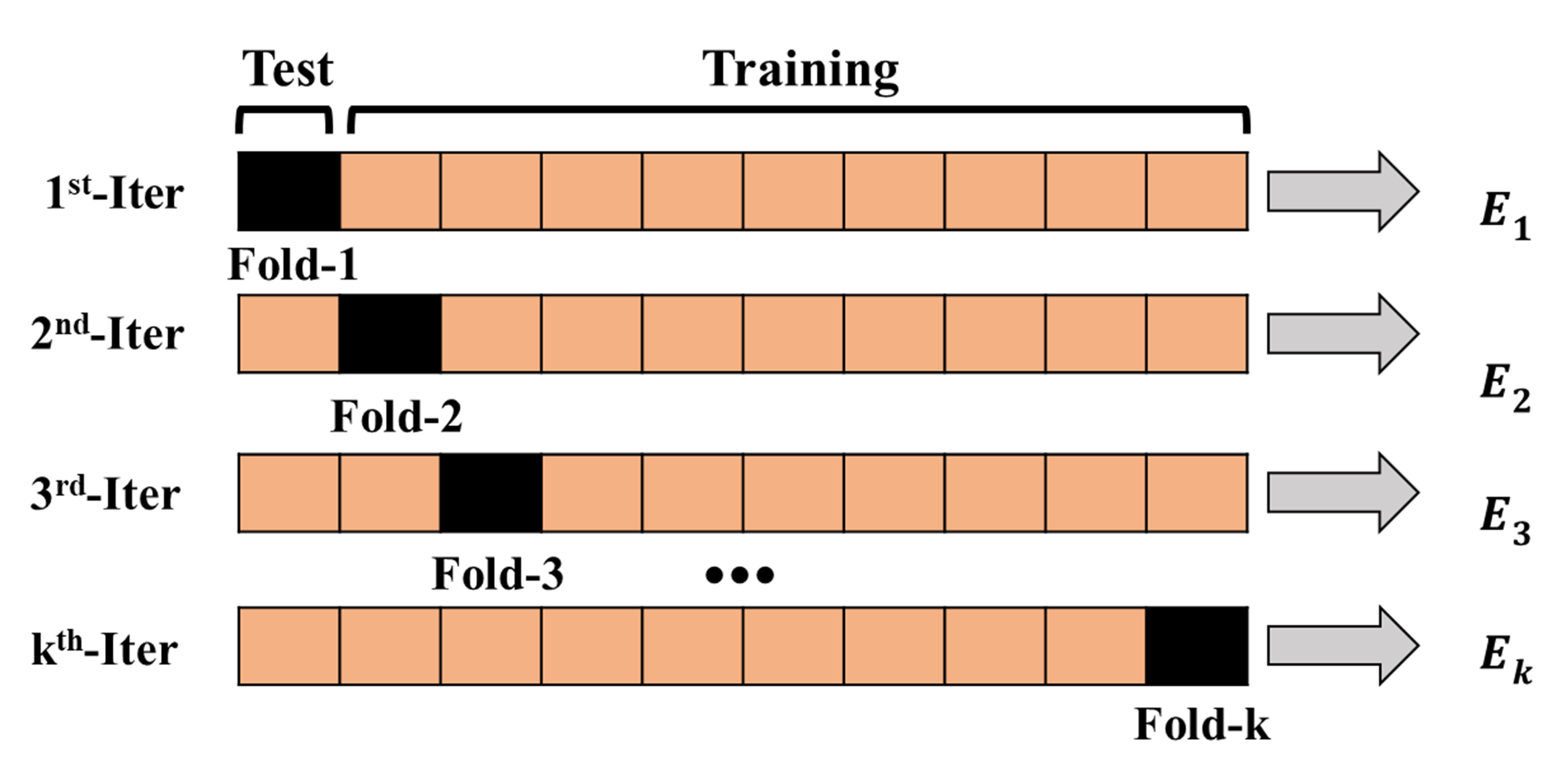}
    \caption{\textit{K-fold cross validation}}
    \label{Fig6}
\end{figure}

The performance of the model with test data can be significantly altered when the proportion of the training set and the test set are changed. In general, random selection with given proportion from an entire data set is separated into test set. However, a test set with another random selection may have different evaluation result especially when there is relatively a few test data. Thus, we strongly recommend using an entire MIT-BIH dataset with \textit{K}-fold cross-validation. \textit{K}-fold cross-validation is a validation technique for stabilizing the performance of the statistical model when the data set is comparatively small. The entire data set is divided into \textit{K} different subsets and repeatedly performing the training with \textit{K}-1 subsets while evaluation is made with a single subset until all K subsets are evaluated. If \textit{K} is set with the number of an entire data set, it is called Leave One-out cross-validation which every ECG data is used once as a test data. In this paper, we set \textit{K} = 10 while the proportion of each data type is near evenly distributed. This method is called Stratified \textit{K}-fold cross-validation. Fig.\ref{Fig6} describes the \textit{K}-fold cross-validation.

\subsubsection{Comparison with two CNN models}
Summarized evaluation results of CNN models are presented in Table \ref{table6} where \textit{Native} refers to training data without the augmentation and \textit{Augmented} refers to with augmentation. The proposed CNN model achieved 0.989 AUC, 99.05\% average accuracy, 99.57\% specificity, 97.85\% average sensitivity and 98.55\% positive predictive value. From the Table \ref{table6}, proposed CNN model with data augmentation shows the best AUC, accuracy, and sensitivity results while AlexNet without augmentation presents the best specificity and positive predictive value. Discrimination of sensitivity between the proposed method and the second highest one is 0.59\% while the discrimination of specificity of the proposed method and the highest one is 0.11\%. This indicates that the proposed model achieved the best accuracy in classifying the arrhythmia while presents similarly the highest accuracy in predicting the normal beat. Interestingly, the VGGNet shows the worst result compared to other two CNN models. The main difference between the VGGNet and the other models is the number of pooling layers. After bypassing four pooling layers, the size of ECG image is reduced to 8 x 8. Although the VGGNet has twice more the number of kernels at the last convolutional layer, it seems that last feature size with 16 x 16 is more stabilized one. Therefore, when using the ECG image size with 128 x 128, we recommend using no more than three pooling layers in CNN model. Between the proposed model and the AlexNet with data augmentation, the difference between the average sensitivities is 0.77\%. This is a quite large difference when we look into detailed sensitivities of the seven different arrhythmias. Table\ref{table7} describes a detailed evaluation of sensitivity with seven types of arrhythmia beats in proposed methods. From the Table\ref{table7}, we can observe that APC, VFW, and VEB arrhythmia classes gain large benefits from the data augmentation which has relatively less ECG data than other arrhythmia classes. Each sensitivity of the APC, VFW, and VEB in the proposed model with data augmentation got 2.89\%, 2.75\%, and 4.72\% higher results than those of second highest results. From these results, we can conclude that data augmentation helps the model to achieve highest average sensitivity with balanced individual sensitivities among seven different arrhythmia types.

\begin{table}[!htb]\small
	\caption{Summarized evaluation results}
    \centering
    \scalebox{1}{
	\begin{tabular}{ llllllll }
    \hline\noalign{\smallskip}
		\textbf{Method} & \textbf{Grade} & \textbf{AUC} & \textbf{Acc(\%)} & \textbf{Sp(\%)} & \textbf{Se(\%)} & \textbf{+P(\%)}\\
        \noalign{\smallskip}\hline\noalign{\smallskip}
		Proposed & Native & 0.986 & 98.90 & 99.64 & 97.20 & 98.63\\
         & Augmented & \textbf{0.989} & \textbf{99.05} & 99.57 & \textbf{97.85} & 98.55\\
        AlexNet & Native & 0.985 & 98.81 & \textbf{99.68} & 96.81 & \textbf{98.63}\\
         & Augmented &0.986 & 98.85 & 99.62 & 97.08 & 98.59\\
        VGGNet & Native & 0.984 & 98.77 & 99.43 & 97.26 & 98.08\\
         & Augmented &0.984 & 98.63 & 99.37 & 96.93 & 97.86\\
         \noalign{\smallskip}\hline
	\end{tabular}}
	\label{table6}
\end{table}

\begin{table}[!htb]\small
	\caption{Detail evaluation in sensitivity}
    \centering
    \scalebox{0.95}{
	\begin{tabular}{ lllllllll }
    \hline\noalign{\smallskip}
		& & & & & \textbf{Se(\%)}\\
        \textbf{Method} & \textbf{Grade}& \textbf{PVC} & \textbf{PAB} & \textbf{RBB} & \textbf{LBB} & \textbf{APC} & \textbf{VFW} & \textbf{VEB}\\
        \noalign{\smallskip}\hline\noalign{\smallskip}
		Proposed & Native & 96.59 & 99.40 & 98.31 & 98.44 & 88.81 & 82.42 & 88.68 \\
         & Augmented & \textbf{97.43} & 99.47 & 98.62 & \textbf{98.54} & \textbf{91.79} & \textbf{90.25} & \textbf{93.40} \\
        AlexNet & Native & 96.25 & 99.10 & 98.12 & 97.85 & 85.95 & 82.75 & 83.55 \\
         & Augmented & 96.58 & \textbf{99.54} & 98.04 & 98.17 & 87.55 & 87.50 & 88.68 \\
        VGGNet & Native & 96.45 & 98.83 & 99.10 & 98.29 & 87.85 & 83.42 & 83.79 \\
         & Augmented & 97.00 & 99.27 & \textbf{99.83} & 98.29 & 88.90 & 86.98 & 87.40 \\
         \noalign{\smallskip}\hline
	\end{tabular}}
	\label{table7}
\end{table}

\begin{figure}[!htb]
	\centering
    \includegraphics[width=\textwidth]{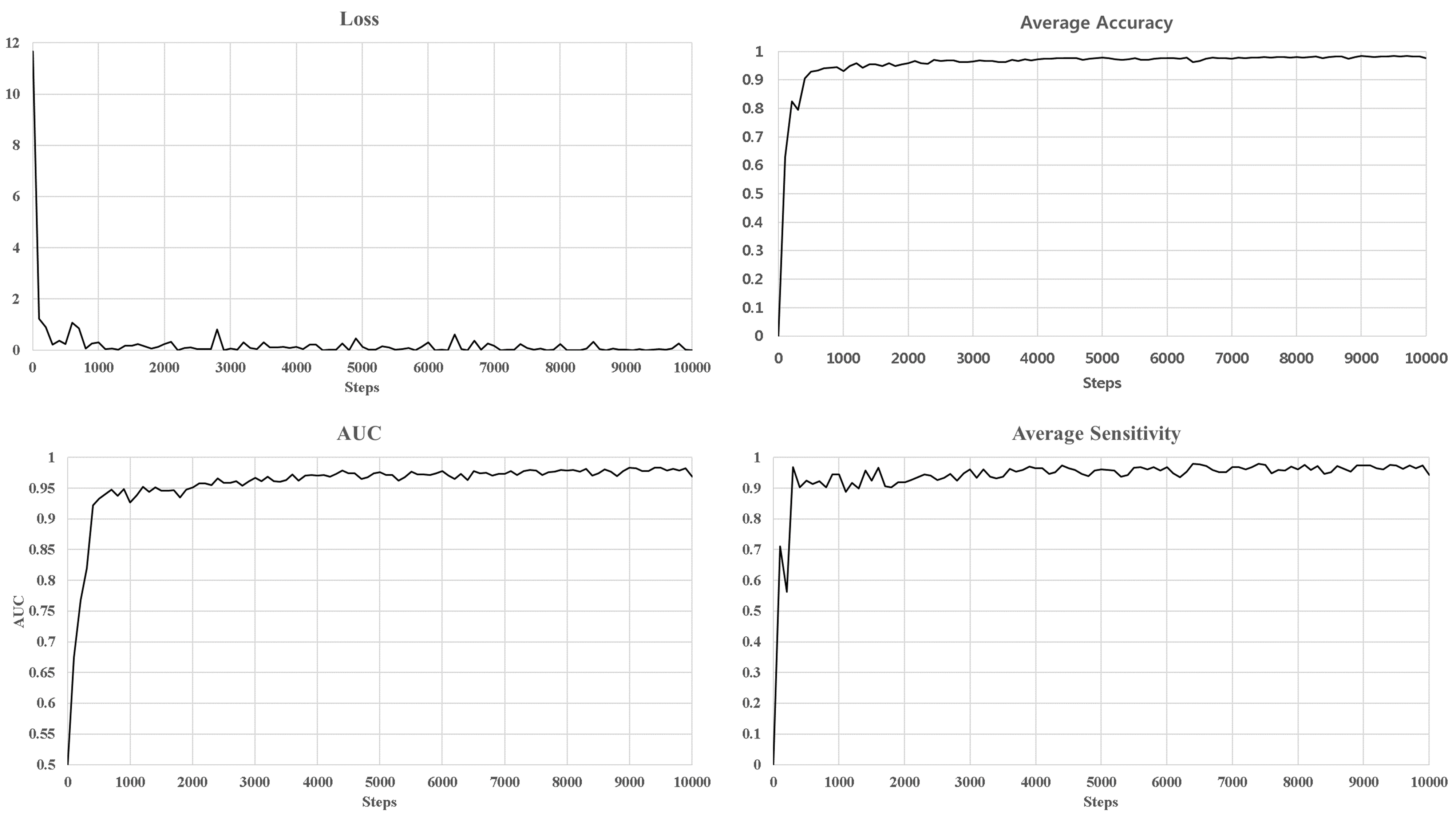}
    \caption{\textit{Evaluation Graphs}}
    \label{Fig7}
\end{figure}

\begin{figure}[!htb]
	\centering
    \includegraphics[width=0.9\textwidth]{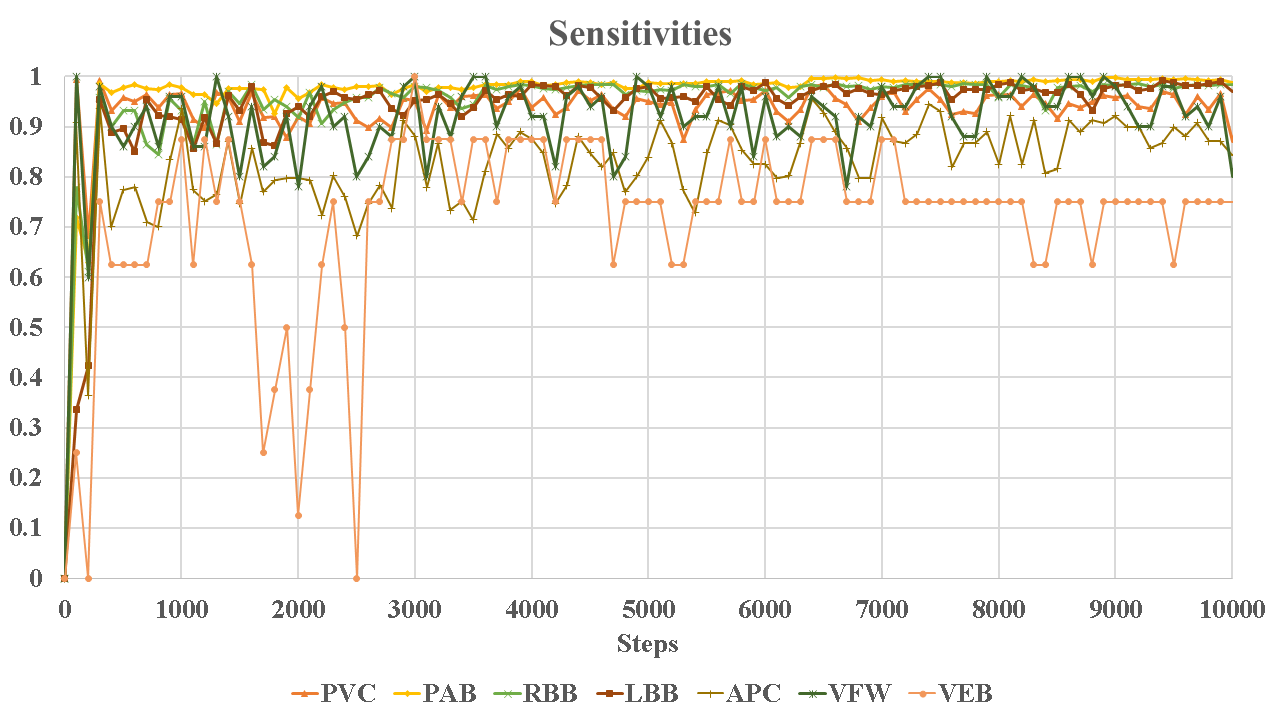}
    \caption{\textit{Each ECG arrhythmia Sensitivity}}
    \label{Fig8}
\end{figure}

The training of proposed CNN model adopts early stopping scheme by using validation set. When the lowest sensitivity does increase for the past 500 global steps, the CNN model stops training and evaluate the model with the test set. Fig.\ref{Fig7} presents graphs of loss, average accuracy, AUC, and average positive predictive value for each global step and Fig.\ref{Fig8} shows individual sensitivity for each arrhythmia type until the 10,000 global steps. From these graphs, loss value starts to converge near 1,000 steps while other evaluation metrics do not. Thus, we can notice that early stopping with loss value may results in low accuracy and sensitivity when there is unbalanced distribution between positive and negative classes. Therefore, we recommend considering other evaluation metrics such as AUC or the average sensitivity when using the early stopping scheme in ECG arrhythmia classification with CNN model.

For more detailed comparison between the proposed CNN, AlexNet, and VGGNet, Table \ref{table8}, Table \ref{table9}, and Table \ref{table10} describe the confusion matrix of each CNN model. The total number of test set may not be equal since some of ECG beats are ignored when randomly splitting into 10 different groups.

\begin{table}[!htb]\small
	\caption{Confusion matrix of proposed CNN model}
    \centering
    \scalebox{1}{
	\begin{tabular}{ lllllllll }
    \hline\noalign{\smallskip}
    	& & & & \textbf{Actual} & & & &\\
		\textbf{Predicted} & NOR & PVC & PAB & RBB & LBB & APC & VFW & VEB \\
        \noalign{\smallskip}\hline\noalign{\smallskip}
		NOR & 74726 & 151 & 2 & 15 & 18 & 110 & 30 & 0 \\
        PVC & 149 & 6947 & 1 & 3 & 7 & 11 & 12 & 0 \\
        PAB & 19 & 13 & 6991 & 2 & 1 & 1 & 1 & 0 \\
        RBB & 83 & 3 & 2 & 7159 & 1 & 9 & 0 & 2 \\
        LBB & 82 & 32 & 1 & 0 & 7957 & 1 & 1 & 1 \\
        APC & 186 & 17 & 0 & 5 & 1 & 2337 & 0 & 0 \\
        VFW & 30 & 15 & 0 & 0 & 1 & 0 & 426 & 0 \\
        VEB & 6 & 1 & 0 & 0 & 0 & 0 & 0 & 99 \\
        \noalign{\smallskip}\hline
	\end{tabular}}
	\label{table8}
\end{table}

\begin{table}[!htb]\small
	\caption{Confusion matrix of AlexNet}
    \centering
    \scalebox{1}{
    \begin{tabular}{ lllllllll }
    \hline\noalign{\smallskip}
    	& & & & \textbf{Actual} & & & &\\
		\textbf{Predicted} & NOR & PVC & PAB & RBB & LBB & APC & VFW & VEB \\
        \noalign{\smallskip}\hline\noalign{\smallskip}
		NOR & 74765 & 131 & 3 & 13 & 30 & 80 & 30 & 1 \\
        PVC & 203 & 6886 & 3 & 1 & 13 & 15 & 9 & 0 \\
        PAB & 14 & 12 & 6996 & 3 & 2 & 1 & 0 & 0 \\
        RBB & 126 & 4 & 1 & 7117 & 3 & 8 & 0 & 0 \\
        LBB & 104 & 39 & 0 & 1 & 7927 & 1 & 3 & 0 \\
        APC & 289 & 14 & 2 & 10 & 2 & 2229 & 0 & 0 \\
        VFW & 42 & 15 & 0 & 0 & 1 & 1 & 413 & 0 \\
        VEB & 10 & 1 & 0 & 0 & 1 & 0 & 0 & 94 \\
        \noalign{\smallskip}\hline
	\end{tabular}}
	\label{table9}
\end{table}

\begin{table}[!htb]\small
	\caption{Confusion matrix of VGGNet}
    \centering
    \scalebox{1}{
	\begin{tabular}{ lllllllll }
    \hline\noalign{\smallskip}
    	& & & & \textbf{Actual} & & & &\\
		\textbf{Predicted} & NOR & PVC & PAB & RBB & LBB & APC & VFW & VEB \\
        \noalign{\smallskip}\hline\noalign{\smallskip}
		NOR & 74784 & 138 & 3 & 15 & 17 & 68 & 27 & 0 \\
        PVC & 200 & 6887 & 3 & 3 & 6 & 9 & 21 & 1 \\
        PAB & 23 & 16 & 6986 & 1 & 0 & 2 & 0 & 0 \\
        RBB & 107 & 7 & 0 & 7136 & 2 & 8 & 0 & 0 \\
        LBB & 92 & 29 & 1 & 0 & 7949 & 1 & 3 & 0 \\
        APC & 264 & 15 & 0 & 5 & 0 & 2261 & 1 & 0 \\
        VFW & 45 & 36 & 1 & 0 & 1 & 0 & 389 & 0 \\
        VEB & 12 & 0 & 0 & 0 & 0 & 0 & 0 & 94 \\
        \noalign{\smallskip}\hline
	\end{tabular}}
	\label{table10}
\end{table}

\subsection{Comparison with existing approaches}
We compared the performance of proposed CNN model with previous ECG arrhythmia classification works. Since these works have a different number of the test set and types of arrhythmia, it is unfair to directly compared with accuracy itself. However, our proposed CNN model achieved successful performance compared to other previous works while introducing the different approach of classifying ECG arrhythmia with two-dimensional images. Table \ref{table11} presents performance comparison with previous works. From the Table \ref{table11}, proposed method achieved the best results in average accuracy, average sensitivity, and average positive predictive value. Narrowing the scope of the number of test set higher than 100,000, proposed method shows the best results in every evaluation metric. The number of arrhythmia types of proposed method is eight while that of Park et al.\cite{park2013arrhythmia} is seventeen since they included other subsidiary heartbeat types from the MIT-BIH arrhythmia database such as nodal escape beat, unclassifiable beat, and non-conducted P wave. However, When considering only the ECG arrhythmia and the normal beat, eight is the maximum number we can get from the MIT-BIH arrhythmia database.

\begin{table*}[!htb]\small
	\caption{Comparison with existing approaches}
    \centering
    \scalebox{0.85}{
	\begin{tabular}{ lllllllll }
    \hline\noalign{\smallskip}
		\textbf{Classifier} & \textbf{Work} & \textbf{\#Type} & \textbf{\#Test set} & \textbf{Acc(\%)} & \textbf{Sp(\%)} & \textbf{Se(\%)} & \textbf{+P(\%)}\\
        \noalign{\smallskip}\hline\noalign{\smallskip}
        \textbf{2D CNN} & \textbf{Proposed} & \textbf{8} & \textbf{106,642} & \textbf{99.05} & 99.57 & \textbf{97.85} & \textbf{98.55}\\
        1D CNN & Kiranyaz et al.\cite{kiranyaz2016real}& 5 & 100,389 & 96.4 & 99.5 & 68.8 & 79.2\\
        FFNN & G\"uler et al.\cite{guler2005ecg}& 4 & 360 & 96.94 & 97.78 & 96.31 & -\\
         & Yu et al.\cite{yu2008integration}& 8 & 4,900 & 98.71 & \textbf{99.65} & - & -\\
        SVM & Melgani and Bazi\cite{melgani2008classification} & 6 & 40,438 & 91.67 & 90.49 & 93.83 & -\\
         & Dutta et al.\cite{dutta2010correlation} & 3 & 93,246 & 95.82 & 99.17 & 86.16 & 97.01\\
        RNN & \"Ubeyli et al.\cite{ubeyli2009combining} & 4 & 360 & 98.06 & 97.78 & 98.15 & -\\
        RFT & Kumar et al.\cite{kumar2012investigating} & 3 & 154 & 92.16 & - & - & -\\
        K-NN & Park et al.\cite{park2013arrhythmia} & 17 & 109,702 & 97 & 95.8 & 96.6 & -\\
        \noalign{\smallskip}\hline
	\end{tabular}}
	\label{table11}
\end{table*}

\section{Discussion and Conclusion}
In this paper, we proposed an effective ECG arrhythmia classification method using two-dimensional convolutional neural networks with ECG image as an input. 128 x 128 grayscale images are transformed from the MIT-BIH arrhythmia database ECG recording. Over 100,000 ECG beat images are obtained with eight types of ECG beats including normal beat and seven arrhythmia beats. Optimized CNN model is designed with considering important concepts such as data augmentation, regularization, and \textit{K}-fold cross-validation. As a result, our proposed scheme achieved 0.989 AUC, 99.05\% average accuracy, 99.57\% specificity, 97.85\% average sensitivity, and 98.55\% average positive predictive value. Our ECG arrhythmia classification result indicates that detection of arrhythmia with ECG images and CNN model can be an effective approach to help the experts to diagnose cardiovascular diseases which can be seen from ECG signals. Furthermore, proposed ECG arrhythmia classification method can be applied to the medical robot or the scanner that can monitors the ECG signals and helps the medical experts to identify ECG arrhythmia more precisely and easily. For the future work, we are building an integrated ECG arrhythmia classification system that scans the patient's ECG monitor through the camera of the medical robot and diagnose the arrhythmia to inform the physician. Proposed ECG arrhythmia classification will be applied the ECG images obtained from the medical robot's camera. 

\begin{acknowledgements}
This research was supported by International Research \& Development Program of the National Research Foundation of Korea(NRF) funded by the Ministry of Science, ICT\&Future Planning of Korea(2016K1A3A7A03952054) and support of Asan Medical Center providing IVUS images and clinical advices for this research are gratefully acknowledged.
\end{acknowledgements}



\end{document}